\documentclass[letterpaper, 10pt, conference]{ieeeconf}  

\IEEEoverridecommandlockouts                                       

\usepackage{graphicx}
\usepackage{amsmath,amssymb}
\usepackage{algorithm}
\usepackage{algorithmicx}
\usepackage{algpseudocode}
\usepackage{array}
\usepackage{url}

\title{\LARGE \bf
Real-time Perceptive Motion Control using Control Barrier Functions with Analytical Smoothing for Six-Wheeled-Telescopic-Legged Robot Tachyon 3
}

\author{
Noriaki Takasugi$^{1}$, Masaya Kinoshita$^{1}$, Yasuhisa Kamikawa$^{1}$, Ryoichi Tsuzaki$^{1}$, Atsushi Sakamoto$^{1}$, \\
$$Toshimitsu Kai$^{1}$, and Yasunori Kawanami$^{1}$
\thanks{$^{1}$Sony Group Corporation, Minato-ku, Tokyo, Japan, 108-0075 {\tt\small noriaki.takasugi@sony.com}}%
}

\begin{document}

\maketitle
\thispagestyle{empty}
\pagestyle{empty}

\begin{abstract}

To achieve safe legged locomotion, it is crucial to generate motion in real-time considering various constraints in robots and environments. In this study, we propose a lightweight real-time perceptive motion control system for the newly developed six-wheeled-telescopic-legged robot, Tachyon 3. In the proposed method, analytically smoothed constraints including Smooth Separating Axis Theorem (SSAT) as a novel higher order differentiable collision detection for 3D shapes is applied to the Control Barrier Function (CBF). The proposed system integrating the CBF achieves online motion generation in a short control cycle of 1 $\bf{ms}$ that satisfies joint limitations, environmental collision avoidance and safe convex foothold constraints. The efficiency of SSAT is shown from the collision detection time of 1 $\boldsymbol{\mu} \bf{s}$ or less and the CBF constraint computation time for Tachyon 3 of several $\boldsymbol{\mu} \bf{s}$. Furthermore, the effectiveness of the proposed system is verified through the stair-climbing motion, integrating online recognition in a simulation and a real machine.

\end{abstract}

\section{INTRODUCTION}

\begin{figure}[t]
    \centering
    \includegraphics[scale=0.40]{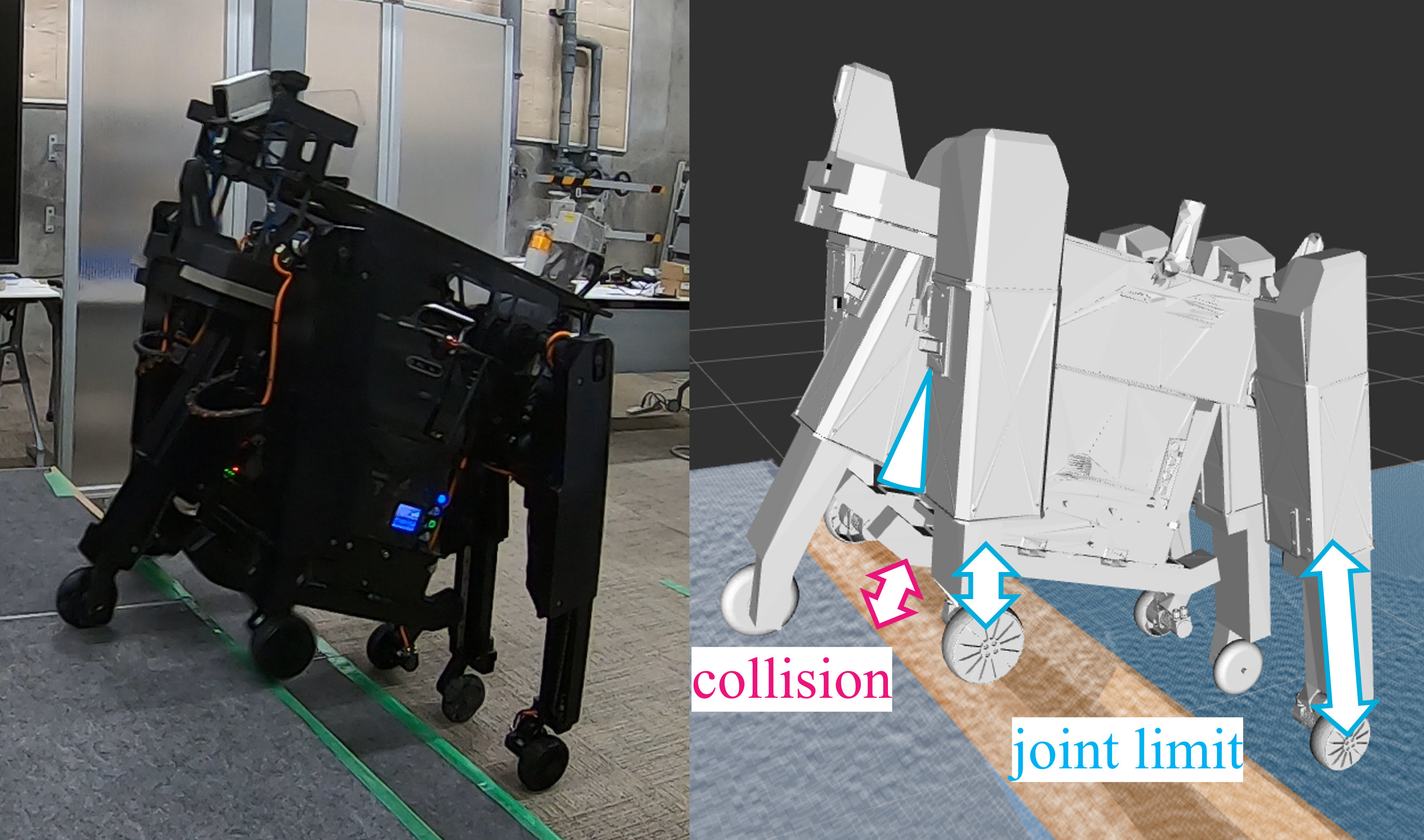}
    \caption{Six-wheeled-telescopic-legged robot Tachyon 3 which has 6 legs and 4 actuated wheels and 2 omni passive wheels.}
  \label{fig:intro}
\end{figure}

In recent years, legged robots have attracted attention from the viewpoint of rough terrain movement performance, and their practical application has advanced. On the other hand, the fall risk remains a challenge for legged robots with a high center of gravity. To mitigate the fall risk of legged robots, Tachyon 3, a novel six-wheeled-telescopic-legged robot, was developed as shown in Fig. \ref{fig:intro} \cite{sonynewmobility}. Tachyon 3 improves stability and energy efficiency through static walking with six legs and wheel movement and lowers the center of gravity with a telescopic leg. However, due to low center of gravity movement, the constraint of avoiding collisions with the environment and the range of motion of the legs becomes severe. Therefore, it is crucial to generate motion under these constraints.

Model Predictive Control (MPC) \cite{bjelonic2021planning} and Reinforcement Learning (RL) \cite{lee2022control} have been actively studied toward motion generation that takes environmental and physical constraints into account. However, due to its high computational complexity, MPC often requires additional high-frequency control for strictly satisfying the constraints as in \cite{bjelonic2021planning}. RL struggles to explicitly handle constraints, making it challenging to learn motion under severe constraints. Control Barrier Function (CBF) \cite{ames2014control} is a more computationally efficient method for considering the constraints, and has been used for real-time collision avoidance between robots and environments \cite{grandia2021multi, liao2023walking, dai2023safe}. However, collision avoidance between convex polyhedron such as cuboid in continuous high-order CBF \cite{nguyen2016exponential, xiao2021high} has not yet been addressed because the high-order derivatives of non-differentiable collision detection are complicated to obtain. Therefore, high-frequency real-time control that takes into consideration collision detection for cuboids has not been achieved.

In this paper, we propose a novel lightweight perceptive real-time control framework for Tachyon 3. Our control methodology operates at a high frequency of 1 kHz, enabling the precise generation of smooth target joint positions for Tachyon 3's actuators, which utilize high-gain position control. We employ the Exponential Control Barrier Function (ECBF) \cite{nguyen2016exponential} to continuously enforce geometric constraints, including the range of motion and environmental collision avoidance. To obtain higher-order derivatives of the collision avoidance for ECBF, we introduce a novel formulation which effectively smoothes the Separating Axis Theorem (SAT) \cite{gottschalk1996separating}. This formulation guarantees collision avoidance between the cuboid body of Tachyon 3 and the surrounding environment. To validate the effectiveness of our proposed method, we conduct practical experiments using both simulation and a real machine. These experiments involve perceptive locomotion in unknown environments, demonstrating the safety and computational efficiency of our approach.

\subsection{Related Works}

Motion generation techniques for legged robots that address environmental constraints can be categorized into two main approaches: model-free and model-based. Model-free motion generation, for instance, has been advanced through perceptive reinforcement learning for both legged robots \cite{miki2022learning} and wheeled-legged robots \cite{lee2022control}. While these methods can generate robust end-to-end motion by utilizing body-relative point clouds for environmental perception, tuning the reward function to learn specific desired motions is challenging, and the assuarance of safety is not guaranteed.

Model-based-control methods, specifically MPC, has also been applied to motion generation of wheeled-legged robots \cite{bjelonic2021planning}. In this prior research, time varying CBFs are employed to satisfy the 2D convex foothold constraints in MPC. There are also studies on legged robots to generate motion by perceptive MPC \cite{grandia2023perceptive}. In \cite{grandia2023perceptive}, a Signed Distance Field (SDF) is constructed from a 3D voxel grid map to avoid collision of swing legs. However, it is limited to collision avoidance between an environment and a point or spherical object, that is, collision avoidance between cuboid are not considered in contrast to this study. Moreover, when using MPC in practice, additional high-frequency low level control is required to continuously satisfy severe constraints in legged locomotion as demonstrated in \cite{bjelonic2021planning, grandia2023perceptive}.

CBF is a fast control method to guarantee safety, such as collision avoidance with the environment. The CBF-based method needs to differentiate the constraints until the control input is explicitly included in the derivative. Collision avoidance constraints between spherical or circular shapes are analytically differentiable by nature, which facilitates their integration with CBFs at any relative order \cite{wang2017safety}. A collision avoidance between a point and a generic environment is typically addressed by using a SDF, with gradients often calculated through finite differences, as often used in MPC. On the other hand, collision detection between cuboids, as examined in this study, is not directly appliable to CBF due to its non-differentiable nature. For collision avoidance between convex shapes, recent advancements include differentiable optimization-based methods \cite{tracy2023differentiable} and Gilbert Johnson Keerthi distance algorithm (GJK)'s randomization \cite{montaut2023differentiable}. The work in \cite{dai2023safe} demonstrates an arm robot achieving collision avoidance through velocity control using CBF with differentiable collision detection. Yet, these techniques are limited to deriving first-order derivatives and are difficult to apply to ECBF which is dealt with in this study. Examples of the direct application of the non-differentiable distance calculation between convex shapes to CBF include a study \cite{thirugnanam2023nonsmooth} using a Non-smooth CBF \cite{glotfelter2017nonsmooth} and a study \cite{liao2023walking} of obstacle avoidance for a legged robot utilizing MPC with discrete-time CBF \cite{zeng2021safety}. Nonetheless, these approaches suffer from increased computational complexity due to the incorporation of dual variables in the optimization process.

As outlined previously, there is a lack of research on the application of continuous high-order CBFs \cite{nguyen2016exponential, xiao2021high} for collision detection between 3D shapes. Consequently, due to computational efficiency, its application to real-time control (e.g., up to 1 kHz) remains unfeasible, and no existing studies have successfully implemented this approach on a real robot with online 3D perception. In \cite{liao2023walking}, although 3D environmental recognition is employed, it is adapted for 2D CBF to circumvent computational complexity. The study \cite{dai2023safe} does not incorporate online obstacle recognition. Furthermore, both \cite{liao2023walking} and \cite{dai2023safe} diverge from our work as they do not utilize the second derivative of collision detection.

\subsection{Contribution}

We introduce a novel perceptive real-time control system for the Tachyon 3. Our methodology encompasses several key innovations. Initially, we present a technique for the smoothing of constraints by substituting non-smooth operations with their analytical smooth operations. Specifically, we introduce a novel collision avoidance CBF by refining the SAT \cite{gottschalk1996separating} through analytical smoothing. This analytical smoothing CBF is versatile, applicable to systems of any relative order, and offers rapid computational performance. The efficacy of the Smooth SAT (SSAT) is substantiated through comparative analysis with alternative collision detection algorithms. Subsequently, we implement a streamlined perceptive real-time control utilizing the SSAT CBF for Tachyon 3. Furthermore, we propose a ground-moving origin-relative model, which simplifies positional control for real-time motion planning. The proposed system is capable of traversing obstacles and stairs. The effectiveness of the proposed method is verified through simulation and real machine integrated online recognition.

\subsection{Outline}

This paper is structured as follows.
In Section \ref{sec:robot}, we explain Tachyon 3's hardware and propose its efficient whole-body kinematic model.
In Section \ref{sec:control}, the perceptive real-time control for Tachyon 3 using ECBF is described.
In Section \ref{sec:t3cbf}, we propose analytically smoothed CBF constraints for Tachyon 3 including SSAT.
In Section \ref{sec:experiments}, we show the effectiveness of our method. Section \ref{sec:conclusions} concludes the paper.

\section{SIX-WHEELED-TELESCOPIC-LEGGED ROBOT TACHYON 3} \label{sec:robot}
\subsection{Hardware Structure}

\begin{figure*}[tb]
    \centering
    \includegraphics[scale=0.07]{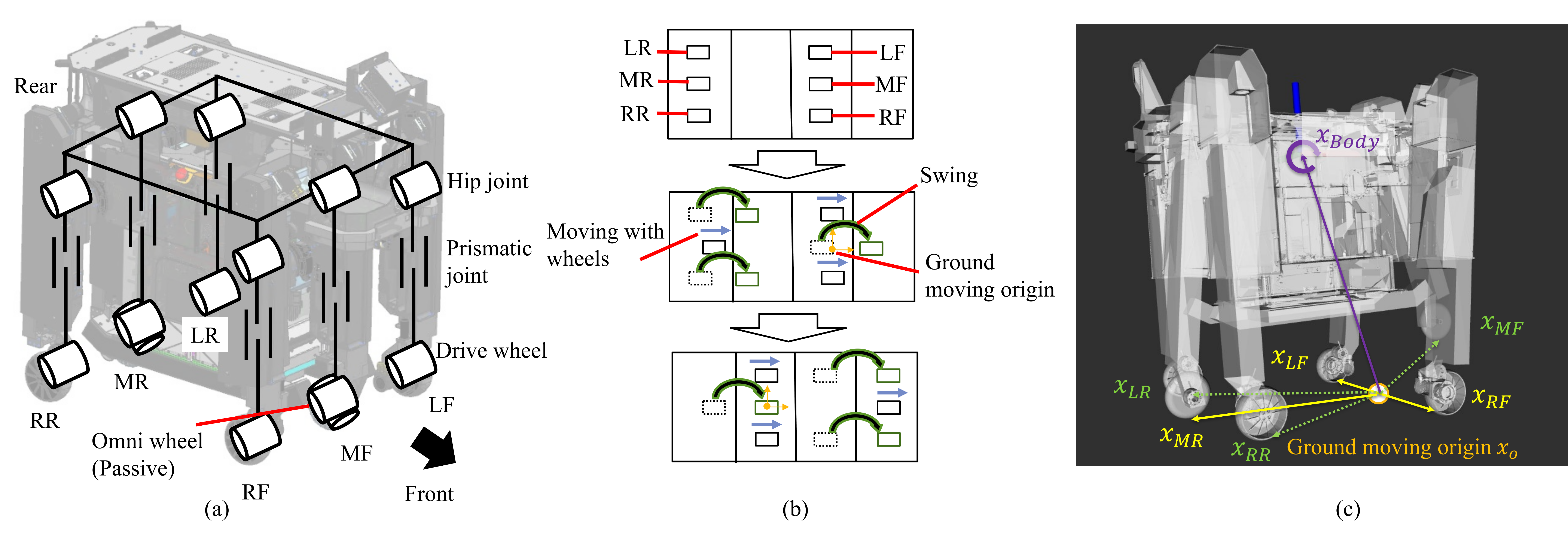}
    \caption{(a) Tachyon 3 is equipped with three front legs (left-front (LF), middle-front (MF), and right-front (RF)) and three rear legs (left-rear (LR), middle-rear (MR) and right-rear (RR)). In total, it consists of 16 active joints including 6 prismatic joints, 6 hip joints and 4 drive wheels, and additional two passive omni wheels. (b) An example of static locomotion for climbing stairs. First, the three wheels, LR, RR and MF, swing to the next steps and the others move with the wheels. Second, MR, LF and RF next steps and the three other wheels move forward. Finally, each set of three wheels is located on the next steps. (c) The kinematic model for CBF consists of the ground moving origin using unicycle dynamics and origin-relative body and end-effector's wheel state.}
  \label{fig:hardware}
\end{figure*}

We developed Tachyon 3, a six-wheeled-telescopic-legged robot with 16 driving and 2 passive joints \cite{sonynewmobility}, improving upon the conventional quadruped Tachyon \cite{kamikawa2021tachyon}.
As shown in Fig. \ref{fig:hardware},
Tachyon 3 features 4 telescopic legs with driving wheels and 2 with omnidirectional passive wheels.
Each leg has 2 degrees of freedom: a hip joint with a 45-degree range of motion (5 degrees inward and 45 degrees outward for the middle legs) and a 500 mm expandable prismatic knee joint. The design of Tachyon 3 addresses technical issues of the six-legged robot \cite{mao2020novel}, reducing leg-stair interference and the robot's footprint. Its low center of gravity enhances stability and fall protection. The robot is equipped with IMU, LiDAR, and 3-axis force sensors on the toes.

The six-wheeled-legged configuration allows stable stair navigation and faster climbing by leveraging wheel movement and maintaining a secure support polygon through the alternating replacement of three support legs. On flat surfaces, Tachyon 3 can achieve free rotation by standing on three legs: the left and right driving wheel legs and the center omnidirectional passive leg without changing steps like wheeled-ANYmal \cite{lee2022control} or steering the drive axis of a wheel.

However, controlling Tachyon 3 on steps with a low center of gravity poses challenges in real-time motion generation while avoiding collisions. We address this using ECBF with a simple wheeled-legged model.

\subsection{Ground Moving Origin Local Model} \label{subsec:model}

Tachyon 3 cannot independently adjust the steering angle of each wheel, which inherently limits the robot's center of rotation to the axle line of either the front or rear drive wheels. To overcome this limitation without creating a complex model, we introduce the concept of a ground moving origin. This concept simplifies the movement of multiple wheels by treating them as a single wheel that shifts its position discretely when the support legs are exchanged. To reduce the impact forces during landing and to decrease lateral sliding during the exchange of support legs, our model includes an acceleration input that smoothly brings the foot's vertical velocity and the body's yaw rate to zero. The robot's state is defined by the relative coordinates of the foot and body with respect to the ground moving origin, which is determined by the wheels in contact with the ground. We assume that the left and right drive wheels move together horizontally to reduce sliding during turns.

The model of the ground moving origin using unicycle dynamics is expressed as:
\begin{equation}
 \label{eq:origin_dynamics}
 \frac{d}{dt}
 \begin{bmatrix}
  p \\
  \psi \\
  \overline{v} \\
  \dot{\psi}
 \end{bmatrix}
 =
 \begin{bmatrix}
  R_z(\psi) \begin{bmatrix}
             \overline{v} \\
             0 \\
             0
            \end{bmatrix} \\
  \dot{\psi} \\
  0 \\
  0
 \end{bmatrix}
 +
 \begin{bmatrix}
  {\mathbf 0} \ {\mathbf 0} \\
  0 \ 0 \\
  1 \ 0 \\
  0 \ 1
 \end{bmatrix}
 \begin{bmatrix}
  \overline{a} \\
  \ddot{\psi}
 \end{bmatrix}
\end{equation}
where $p \in \mathbb{R}^3$ is the position of the wheel in the world coordinate system. $\overline{v} \in \mathbb{R},\ \overline{a} \in \mathbb{R}$ are the velocity and acceleration relative to the wheel, respectively. $\psi \in \mathbb{R}$ is the rotation angle of the wheel. Additionally, the wheel position and body posture relative to the origin is expressed as:
\begin{equation}
 \label{eq:local_dynamics}
 \frac{d}{dt}
 \begin{bmatrix}
  q \\
  \dot{q}
 \end{bmatrix}
 =
 \begin{bmatrix}
  \dot{q} \\
  0
 \end{bmatrix}
 +
 \begin{bmatrix}
  0 \\
  1
 \end{bmatrix}
 \ddot{q}
\end{equation}
where $q, \dot{q}, \ddot{q}$ represent origin-relative value, velocity, and acceleration, respectively.

Based on equations (\ref{eq:origin_dynamics}) and (\ref{eq:local_dynamics}), the whole-body kinematic model of Tachyon 3 can be described as shown in Fig. \ref{fig:hardware} (d):
\begin{gather}
 \label{eq:dynamics}
 \frac{dx}{dt} = f(x) + g(x)u  \\
 x =
 \begin{bmatrix}
  x_{\rm o} & x_{\rm body} & x_{\rm ee_1} & \cdots & x_{\rm ee_6}
 \end{bmatrix}^T
 \nonumber \\
 u =
 \begin{bmatrix}
  u_{\rm o} & u_{\rm body} & u_{\rm ee_1} & \cdots & u_{\rm ee_6}
 \end{bmatrix}^T
 \nonumber \\
 f =
 \begin{bmatrix}
  f_{\rm o} & f_{\rm body} & f_{\rm ee_1} & \cdots & f_{\rm ee_6}
 \end{bmatrix}^T
 \nonumber \\
 g =
  \begin{bmatrix}
  g_{\rm o} & g_{\rm body} & g_{\rm ee_1} & \cdots & g_{\rm ee_6}
 \end{bmatrix}^T
 \nonumber
\end{gather}
In the formulation, $x_{\rm o} \in \mathbb{R}^6$ and $u_{\rm o} \in \mathbb{R}^2$ represent the state and input, respectively, of the ground moving origin dynamics as defined in Equation (\ref{eq:origin_dynamics}). Similarly, $x_{\rm body} \in \mathbb{R}^6$ and $u_{\rm body} \in \mathbb{R}^3$ denote the origin-relative body's x, z and pitch state and input of the linear dynamics as defined in Equation (\ref{eq:local_dynamics}). Additionally, $x_{\rm ee_i} \in \mathbb{R}^4$ and $u_{\rm ee_i} \in \mathbb{R}^2$ represent the origin-relative end-effector's x and z state and input of the linear dynamics as defined in Equation (\ref{eq:local_dynamics}).
The total number of states $N_x \in \mathbb{N}$ is 36, and the total number of inputs $N_u \in \mathbb{N}$ is 17.

This model is computationally efficient because it combines multiple nonholonomic constraints into a single equation. In addition, since the coordinate system is relative to the ground, it is easy to set the unilateral contact constraint of the wheel and to handle position control.
However, this kinematic model doesn't account for full dynamics, which may be crucial during rapid movements.
Future work should incorporate a more detailed dynamic model into the CBF framework to enhance safety across various conditions.

\section{REAL-TIME PERCEPTIVE MOTION CONTROL FOR TACHYON 3} \label{sec:control}

\begin{figure}[tb]
    \centering
    \includegraphics[scale=0.07]{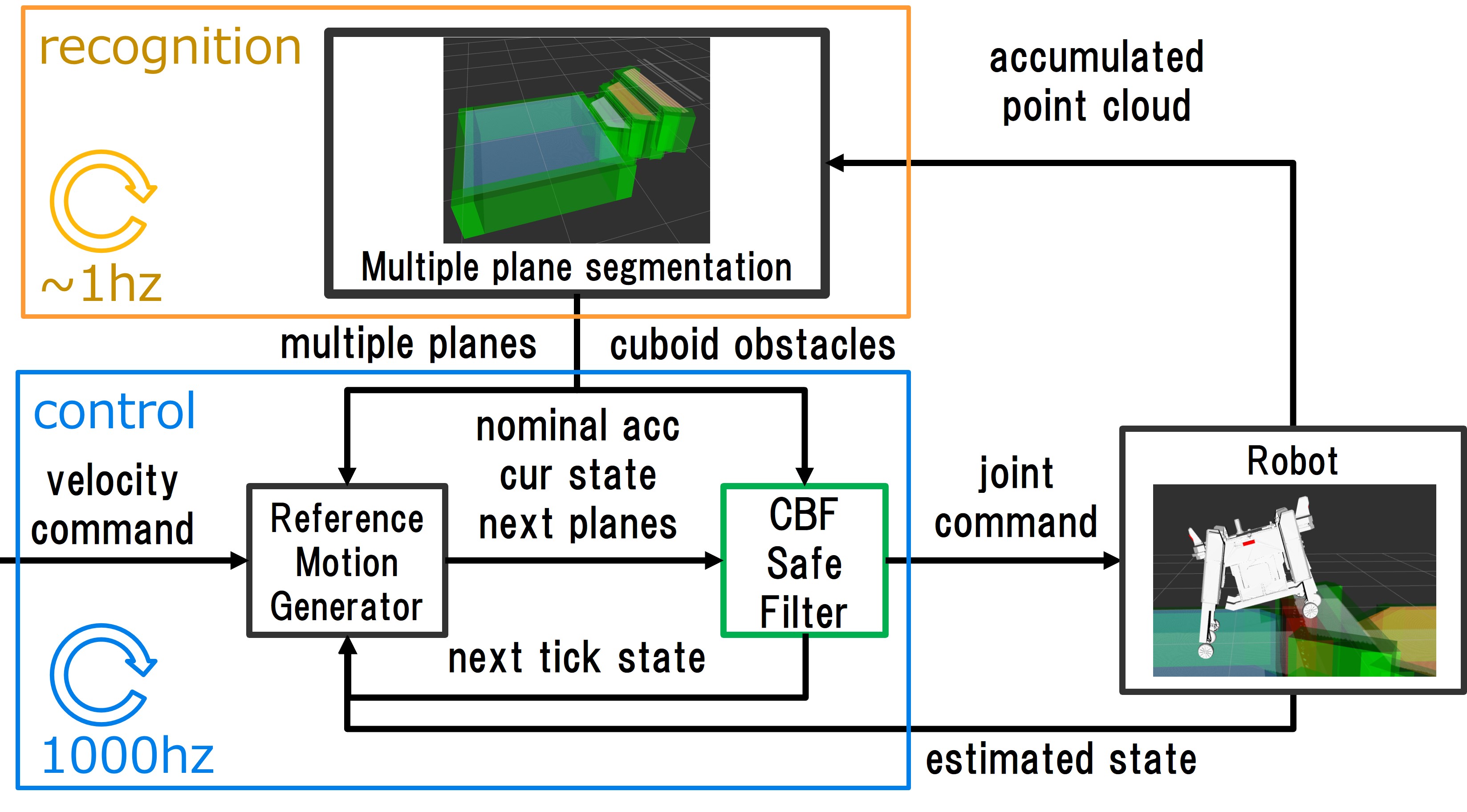}
    \caption{Overview of real-time perceptive motion control for the six-wheeled-telescopic-legged robot Tachyon 3.}
  \label{fig:system}
\end{figure}

This chapter describes a perceptive real-time control system utilizing ECBF for Tachyon 3. Fig. \ref{fig:system} shows an overview of the proposed system. The recognition process runs at 1 Hz and the control process runs at 1000 Hz. The proposed system receives the velocity command from the user and outputs the joint command for the next control cycle taking the percieved environmental information. Then, it is converted to the motor current of the joint through PD control.
In the following, first, we present an outline of recognition, and then reference motion generation is described. Finally, safety filter using ECBF for Tachyon 3 is shown.

\subsection{Recognition}

In this study, planes are extracted by multiple plane segmentation from accumulated point clouds obtained by integrating point clouds from LiDAR and an estimated state. In multiple plane segmentation, clusters of point clouds are generated by region growing \cite{rabbani2006segmentation}, and polygonal planes are extracted from each cluster point cloud using RANSAC algorithm \cite{schnabel2007efficient} and the convex hull algorithm. In addition, a cuboid is extracted from the polygonal plane along the edge to avoid collision with the edge of the step. These polygonal planes and cuboid obstacles are used in real-time motion control.

\subsection{Reference Motion Generation}

In reference motion generation, the reference footstep and wheel speed in 2D are calculated from the velocity command. As described in Section \ref{sec:robot}, the Tachyon 3 requires a zero yaw rate when the front and rear driving wheels are landed due to its structure. Therefore, a body's yaw trajectory is generated by cubic spline interpolation so that the yaw rate becomes zero when the front and rear driving wheels are landed in response to the turn command. Subsequently, these 2D references are modified to 3D utilizing multiple planes. For 3D adjustment, the nearest plane from the nominal footstep is selected from multiple planes, and the footstep is projected on the selected plane. Since the extracted plane can become a non-convex region when obstacles are included, a safe convex foothold is extracted through a method similar to \cite{grandia2023perceptive}. Furthermore, to mitigate unnecessary stepping, ground contact is maintained when the wheels traverse linearly along the same plane.

\subsection{CBF QP Formulation} \label{subsec:cbfqp}

We employ the ECBF \cite{nguyen2016exponential} to address the high relative degree constraints arising from position limitations with acceleration inputs for the Tachyon 3's kinematic model (\ref{eq:dynamics}). To impose ECBF constraints on the nominal input $u^{ref} \in \mathbb{R}^{N_u}$, the following convex QP is solved.
\begin{align}
 \label{eq:cbfqp}
  u(x) = \mathop{\arg\min}_{u \in \mathbb{R}^{N_u},\delta \in \mathbb{R}^{N_{CBF}}} {
 \left\| u - u^{ref} \right\|^2_{w_u} + \left\|\delta \right\|^2_{w_{\delta}}
 } \\
 s.t \ L_f^{r_i}h_i(x) + L_gL_f^{r_i - 1}h_i(x)u \geq -K_{i}\eta_{i}(x) + \delta_i \nonumber \\
  \eta_{i}(x) =
   \begin{bmatrix}
    h_i(x) \\
    \dot{h_i}(x)  \\
    \vdots  \\
    h_i^{r_i-1}(x))
   \end{bmatrix}
   =
   \begin{bmatrix}
    h_i(x) \\
    L_f h_i(x) \\
    \vdots \\
    L_f^{r_i-1} h_i (x)
   \end{bmatrix}
   \nonumber \\
 A u \leq b \nonumber \\
 i = 1, ..., N_{cbf} \nonumber
\end{align}
Here, $w_u \in \mathbb{R}^{N_u}$ weights nominal input tracking, $h \in \mathbb{R}^{N_{CBF}}$ denotes ECBF constraints, and $L_f$, $L_g$ are Lie derivatives for $f$, $g$ in (\ref{eq:dynamics}). $r_i \in \mathbb{N}$ is the relative degree, with $K_i \in \mathbb{R}^{r_i}$ and $\eta_i \in \mathbb{R}^{r_i}$ representing ECBF gain and time derivatives of $h_i \in \mathbb{R}$. $K_i$ are manually selected to minimize input modifications while aiming for feasibility. Additional input constraints, including control limits, are defined by $A \in \mathbb{R}^{N_c \times N_u}$ and $b \in \mathbb{R}^{N_c}$. For infeasible cases where no solution satisfies both additional and ECBF constraints, we introduce a relaxation parameter $\delta \in \mathbb{R}^{N_{CBF}}$ with penalty weight $w_{\delta} \in \mathbb{R}^{N_{CBF}}$ to compute approximate solutions. This slack variable allows minor violations of pre-margined constraints, designed to absorb modeling errors and uncertainties, such as those in cuboid obstacles avoidance.

Target inputs are integrated via (\ref{eq:dynamics}) to obtain wheel velocities and body-relative foot positions. These foot positions are then converted to leg joint angles using inverse kinematics.

\section{ANALYTICAL SMOOTH CBF CONSTRAINTS FOR TACHYON 3} \label{sec:t3cbf}

\begin{figure}[tb]
    \centering
    \includegraphics[scale=0.13]{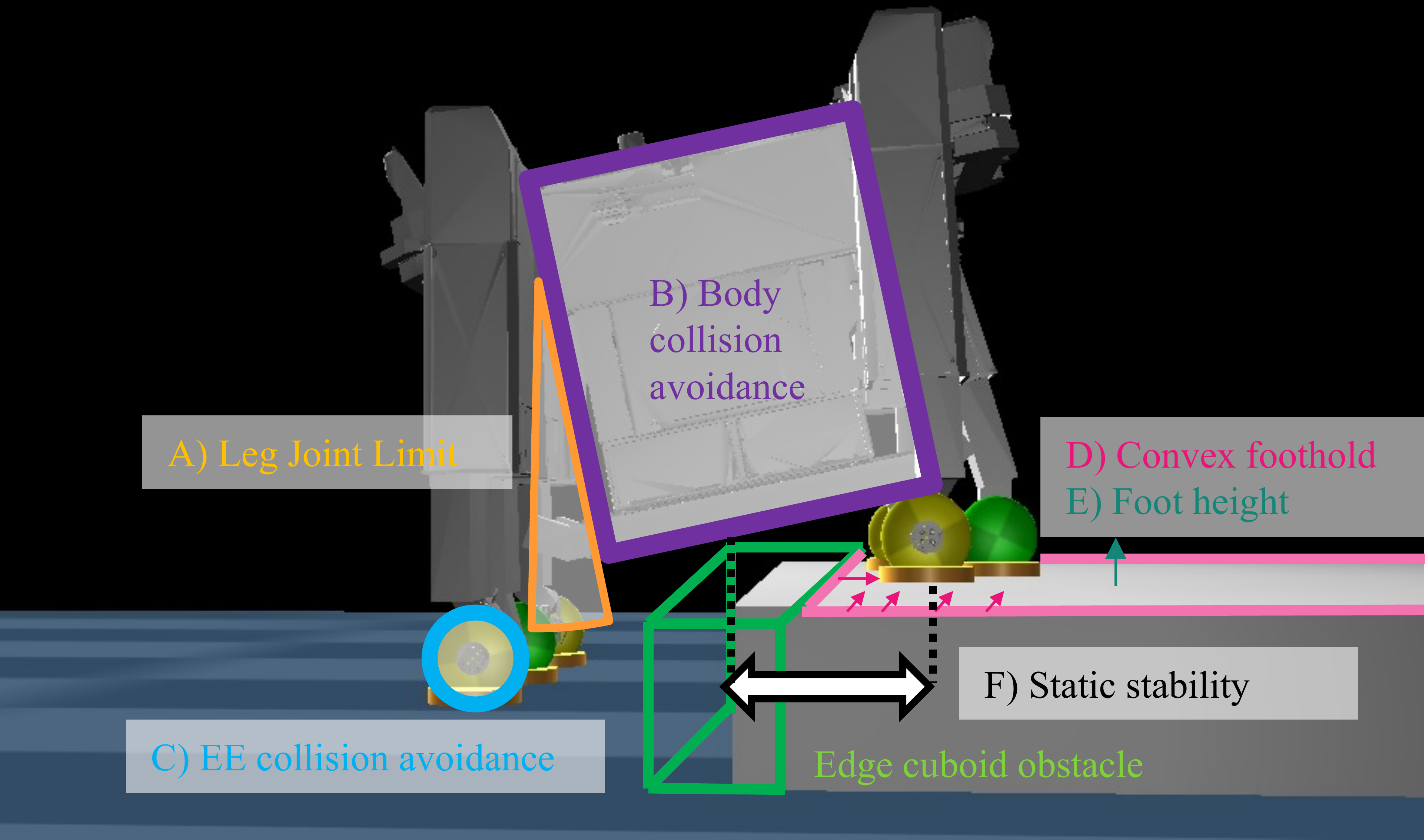}
    \caption{The six types of ECBF for Tachyon 3's geometric constraints.}
  \label{fig:t3cbf}
\end{figure}

This chapter describes how to design continuously differentiable constraints of ECBF for Tachyon 3's safe locomotion as shown in Fig. \ref{fig:t3cbf}. Specifically, non-smooth operations in constraints are replaced to smooth approximated operations. We believe that the present method, particulary the smoothing techniques in CBFs, are applicable and useful to widely types of robots.

\subsection{Leg joint limit}

The leg joint angle limits are defined by converting from the relative position of each leg to the joint angle as follows because the Tachyon 3's knee joint is a prismatic joint.
\begin{align}
 y_1(x) = y_{\rm knee} = \sqrt{\overline{p}_x^2 + \overline{p}_y^2},\ \ y_2(x) = y_{\rm hip} = {\rm atan2}(\overline{y}, \overline{x}) \nonumber
\end{align}
where $\overline{p}_x, \overline{p}_y \in \mathbb{R}$ are the relative toe position from the base of the leg. $y_{knee}, y_{hip} \in \mathbb{R}$ represents the length of knee joint and the angle of hip joint, respectively. The min / max limits on the joint angles of each leg are described as follows.
\begin{align}
 h_{i, {\rm MIN}}(x) &= y_i(x) - y_{\rm MIN} \\
 h_{i, {\rm MAX}}(x) &= y_{\rm MAX} - y_i(x)
\end{align}

\subsection{Body collision avoidance constraints}

We impose collision avoidance constraints between the body of Tachyon 3 and the edges of the multiple plane environments. Both the plane edges and Tachyon 3's body are represented as cuboids. To construct differentiable CBFs for ensuring collision avoidance, the SAT \cite{gottschalk1996separating} is utilized to efficiently detect collision between two cuboids. SAT checks whether there is any overlap along the candidate separating axes and determines that there is no collision if there is no overlap in all the separating axes. For simple geometric shapes like cuboids, the SAT is efficient because there are only a few potential separating axes, and specifically, it is computationally more efficient than the popular GJK \cite{gilbert1988fast}, an efficient collision detecting method between complex convex polytopes.

\begin{figure}[tb]
    \centering
    \includegraphics[scale=0.4]{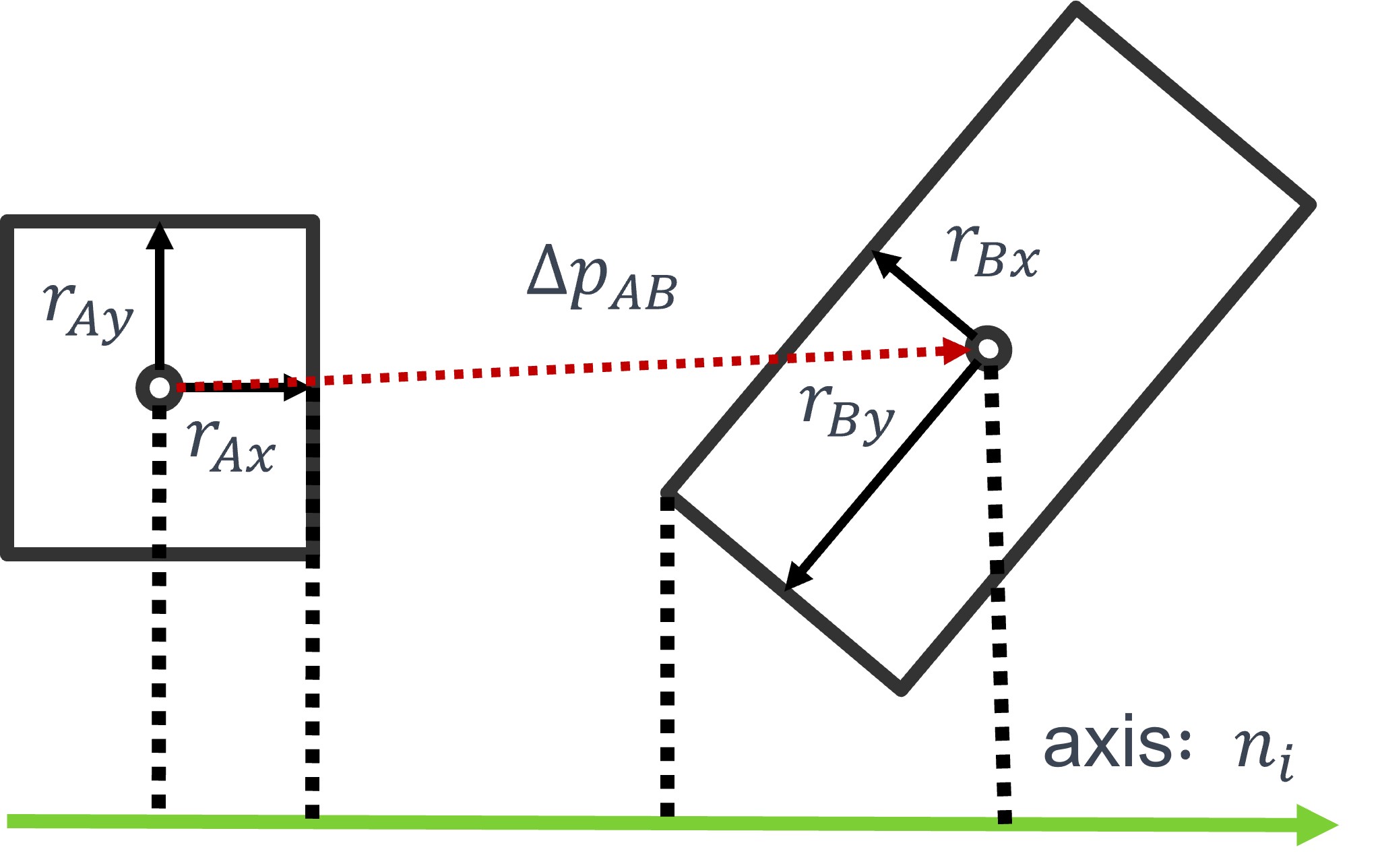}
    \caption{Separating axis theorem for collision detection between two oriented bounding boxes (2D).}
  \label{fig:sat}
\end{figure}

In this study, we present a novel approach to smooth the SAT, which is typically non-differentiable, by using analytical smoothing functions. For better understanding, we provide a 2D example of SAT in Fig. \ref{fig:sat}. There are four potential axes, each corresponding to the direction of a side. Extending this to 3D, we have nine axes from the cross products of each cuboid face's normals, and six axes along the normals of the faces. The collision detection method for the 3D case, using these axis candidates, is described as follows.
\begin{align}
 \label{eq:sat}
 h_{A,B} &= \max(y_1(x),...,y_{15}(x))) \\
 y_{i} &= |n_i \cdot \Delta p_{AB}| - \sum_{\substack{j=A,B,\\k=x,y,z}}|n_i \cdot r_{jk}| \nonumber
\end{align}
where $n_i \in \mathbb{R}^3$ is the separating axis, $\Delta p_{AB} \in \mathbb{R}^3$ is a vector connecting the centers of two cuboids, and $r_{jk} \in \mathbb{R}^3$ is a vector representing the dimensions (vertical, horizontal, and height) of the cuboid.
(\ref{eq:sat}) can be smoothed by replacing the non-smooth maximum and absolute value function.

First, the Log Sum Exp (LSE) \cite{blanchard2021accurately} is adopted as an approximation to the maximum value function with bounds as in \cite{molnar2023composing},
\begin{align}
 \label{eq:lse}
 f_{\rm smax}(x_1, ... ,x_n) = \frac{1}{\alpha} {\rm log} \sum_{i = 1}^n {\rm exp}(\alpha x_i)  \\
 0 \leq f_{\rm smax}(x_1, ..., x_n) - \max(x_1, ..., x_n) \leq \frac{\rm log(n)}{\alpha} \nonumber
\end{align}
Here, the parameter $\alpha \in \mathbb{R}$ modulates the degree of smoothness, with larger values yielding greater accuracy.
To address the overflow problem associated with LSE, we apply it only when the output value is smaller than a threshold, using the calculation of the non-differentiable SAT.

Next, the xtanh function is employed as a precise surrogate for the absolute value function, as described in the following expression \cite{bagul2017smooth}.
\begin{align}
 \label{eq:xtanh}
 f_{\rm sabs}(x) = x{\rm tanh}(\alpha x)  \nonumber \\
 |f_{\rm sabs}(x) - |x|| < \frac{1}{\alpha}
\end{align}
The parameter $\alpha \in \mathbb{R}$ adjusts the smoothness of the approximation, similar to LSE.

Finally, the detection of collision between the body and the edge is achieved by smoothing (\ref{eq:sat}) using (\ref{eq:lse}) and (\ref{eq:xtanh}): 
\begin{align}
 h_{A,B} &= f_{\rm smax}(y_1(x),...,y_{15}(x))) \\
 y_{i}(x) &= f_{\rm sabs}\left( n_i \cdot \Delta p_{A,B} \right) - \sum_{\substack{j=A,B,\\k=x,y,z}} f_{\rm sabs}(n_i \cdot r_{jk})  \nonumber
\end{align}

\subsection{End efector collision avoidance constraints}

The toe wheel is approximated by a sphere. The smooth function for detection of collision between the sphere toe and the cuboid edge is expressed using superellipsoid:
\begin{align}
 h(x) &= \left(\frac{\overline{p}_X(x)}{a}\right)^{2N} + \left(\frac{\overline{p}_Y(x)}{b}\right)^{2N} + \left(\frac{\overline{p}_Z(x)}{c}\right)^{2N} - 1 \\
 \overline{p}(x) &= R^T_{\rm edge}(p_{\rm foot}(x) - p_{\rm edge}) \nonumber
\end{align}
where $\overline{p}(x) \in \mathbb{R}^3$ is the relative position of the toe from the center of the cuboid $p_{edge} \in \mathbb{R}^3, R_{edge} \in \mathbb{R}^{3\times3}$. $a, b, c \in \mathbb{R}$ is a parameter representing the length(x), width(y), and height(z) of the cuboid, respectively, where, $N \in \mathbb{N}$ is a parameter used to adjust the smoothing. A larger value of $N$ leads to a more precise cuboid, while a smaller value of $N$ results in smoother behavior at non-differentiable vertices.

\subsection{Convex foothold constraints}

The convex foothold constraints are applied to CBF.
\begin{equation}
 h(x) = a^T_{ij}p_i(x) + b_{ij}
\end{equation}
where $a_{ij} \in \mathbb{R}^2, b_{ij} \in \mathbb{R}$ represent the half-space constraints of the $j$ th edge of the convex polygon for the world toe xy position $p_{i}(x) \in \mathbb{R}^2$'s safe region. These constraints are activated during the foot-lowering phase of the swing and remain in effect throughout the entire stance phase. When the leg is within the safe convex region, the CBF ensures it does not step outside this region. Conversely, when the swing leg is outside the convex safe region, it functions as a Control Lyapunov Function, guiding the leg towards the safe region, as demonstrated in \cite{xiao2021high}.

\subsection{Foot height constraints}
To prevent collisions with the horizontal plane, a height constraint of $h(\overline{z}) = \overline{z}_i - \overline{z}_{floor}$ is applied to each toe. This ensures each toe remains above the floor level.

\subsection{Approximate static stability constraints}
To maintain static stability, we constrain the body position within the convex hull of the feet. This serves as a proxy for keeping the center of mass in the support polygon. The constraint regarding the local translational position between the body $\overline{p}_{body} \in \mathbb{R}$ and the toes $\overline{p}_{i} \in \mathbb{R}$ is formulated as:
\begin{equation}
 h(x) = a_i(\overline{p}_i(x) - \overline{p}_{body}(x)) + b_i
\end{equation}
For each foot, both upper and lower limit constraints are applied, resulting in two constraints per foot to define the boundaries of the convex hull.


\section{EXPERIMENTS} \label{sec:experiments}
We herein show the effectiveness of the proposed method through experiments. First, we present a benchmark for the computation time of the proposed SSAT against the previous method. Then, the effectiveness of real-time control for Tachyon 3 using SSAT CBF was confirmed in a simulation. Finally, the proposed method was integrated on the real machine.

\subsection{Comparison of Smooth Collision Detection}
\begin{table}[tb]
 \caption{AVERAGE COMPUTATION TIMES OF CUBOID COLLISION DETECTION}
 \label{tb:cputime}
 \centering
 \begin{tabular}{|m{12mm}|m{7mm}|m{7mm}|m{10mm}|m{6mm}|m{6mm}|m{6mm}|} \hline
  Method & SSAT with \par LSE and xtanh & SSAT with \par LSE and sqrt & SSAT with \par Boltz and xtanh & SAT & GJK & LP \\ \hline
  evaluate \par [$\mu$s]  & 0.47 & 0.57  & 0.47  & 0.05 & 0.50 & 9.35 \\ \hline
  evaluate + \par differentiate [$\mu$s]& 0.78 & 0.93 & 1.04 & - & - & - \\ \hline
 \end{tabular}
\end{table}
\begin{figure}[tb]
    \centering
    \includegraphics[scale=0.175]{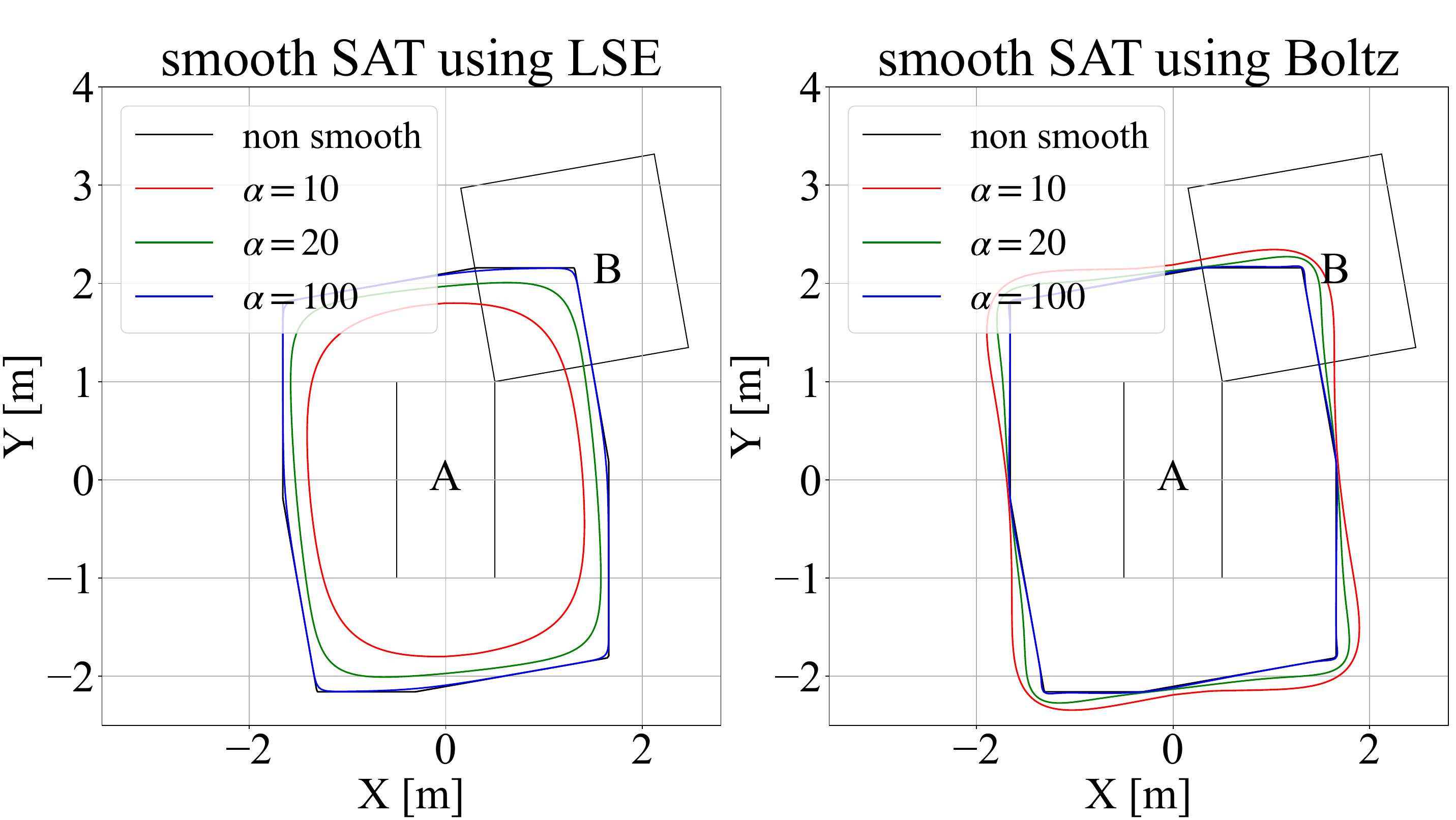}
    \caption{The collision detection boundary using the smooth maximum method for 2D SAT. The left graph uses LSE, while the right graph uses the Boltzmann operator. The red, green, and blue lines represent the trajectory of the center point of box B as it comes into contact with box A, with different values of the smoothing parameter $\alpha$. Box B is movable on the XY plane and tilted at a 10-degree angle. The black line represents the boundary when the SAT is used for collision detection.}
  \label{fig:smooth_sat_comparison}
\end{figure}

The computational efficiency of SSAT is verified by comparing with other smoothing functions and other collision detection methods. Table \ref{tb:cputime} shows the results of the calculation time of random cuboid's collision detection. SSAT with LSE and xtanh is the proposed method explained in Section \ref{sec:t3cbf}. For comparison, two other smoothing functions are considered for SAT : the square root of square (sqrt) $f_{sabs}(x) = \sqrt{(x + \epsilon)^2}$ with a smoothing parameter $\epsilon \in \mathbb{R}$ as smooth absolute value function, and the Boltzmann operator (boltz) $ f_{smax}(x_1, ... ,x_n) = \frac{\sum_{i=1}^n x_i exp(\alpha x_i)}{\sum_{i=1}^n exp(\alpha x_i)}$ with smoothing parameter $\alpha \in \mathbb{R}$ as smooth miximum. SAT and GJK are non-differentiable implementation in hpp-fcl \cite{panhpp}. LP refers to the previous method \cite{dai2023safe, tracy2023differentiable} using Linear Programming.

While the proposed SSAT is slower than the non-differentiable SAT due to the additional smoothing process, it is sufficiently fast to be comparable to GJK (within 1 $\mu$s). Furthermore, compared with the optimization-based method, the proposed method is more than 10 times faster. Therefore, the proposed method is the fastest differentiable cuboid collision detection.
About SSAT, the computation time for calculating collision detection value, it's Jacobian and Hessian with respect to the one box's position was also measured. The result of Table \ref{tb:cputime} indicates that utilizing LSE and xtanh is most efficient for SSAT.

The effectiveness of the SSAT was assessed by employing both the LSE and the Boltzmann operator as potential smooth maximum functions. As depicted in Fig. \ref{fig:smooth_sat_comparison}, the Boltzmann operator results in a non-convex shape, whereas the LSE maintains a convex shape, making it easier to handle within collision avoidance. Furthermore, the LSE has constant upper and lower bounds for the error of the maximum function (\ref{eq:lse}). By incorporating a margin from (\ref{eq:lse}), (\ref{eq:xtanh}), it is possible to strictly satisfy collision avoidance. Considering these advantages and the computational time aspect mentioned earlier, the LSE has been selected for SSAT.

Here we examined the computation time of a single CBF constraint for SSAT of Tachyon 3. Since the proposed method can be computed analytically, code generation \cite{vcertik2013symengine} using symbolic differentiation is performed. Thanks to the offline optimization of code, sparse data can be handled efficiently. Furthermore, constant computational complexity is suitable for achieve real-time control. On an embedded PC of Tachyon 3, the computational time of CBF calculation was less than 4 $\mu$s. This computation time is faster than the collision detection of LP in Table \ref{tb:cputime}, even though the information of the second derivative is required. As such, it can be a faster method than the existing CBF \cite{liao2023walking, dai2023safe} using optimization-based method.

\subsection{Simulation Experiments for Tachyon 3}
\begin{figure}[tb]
    \centering
    \includegraphics[scale=0.13]{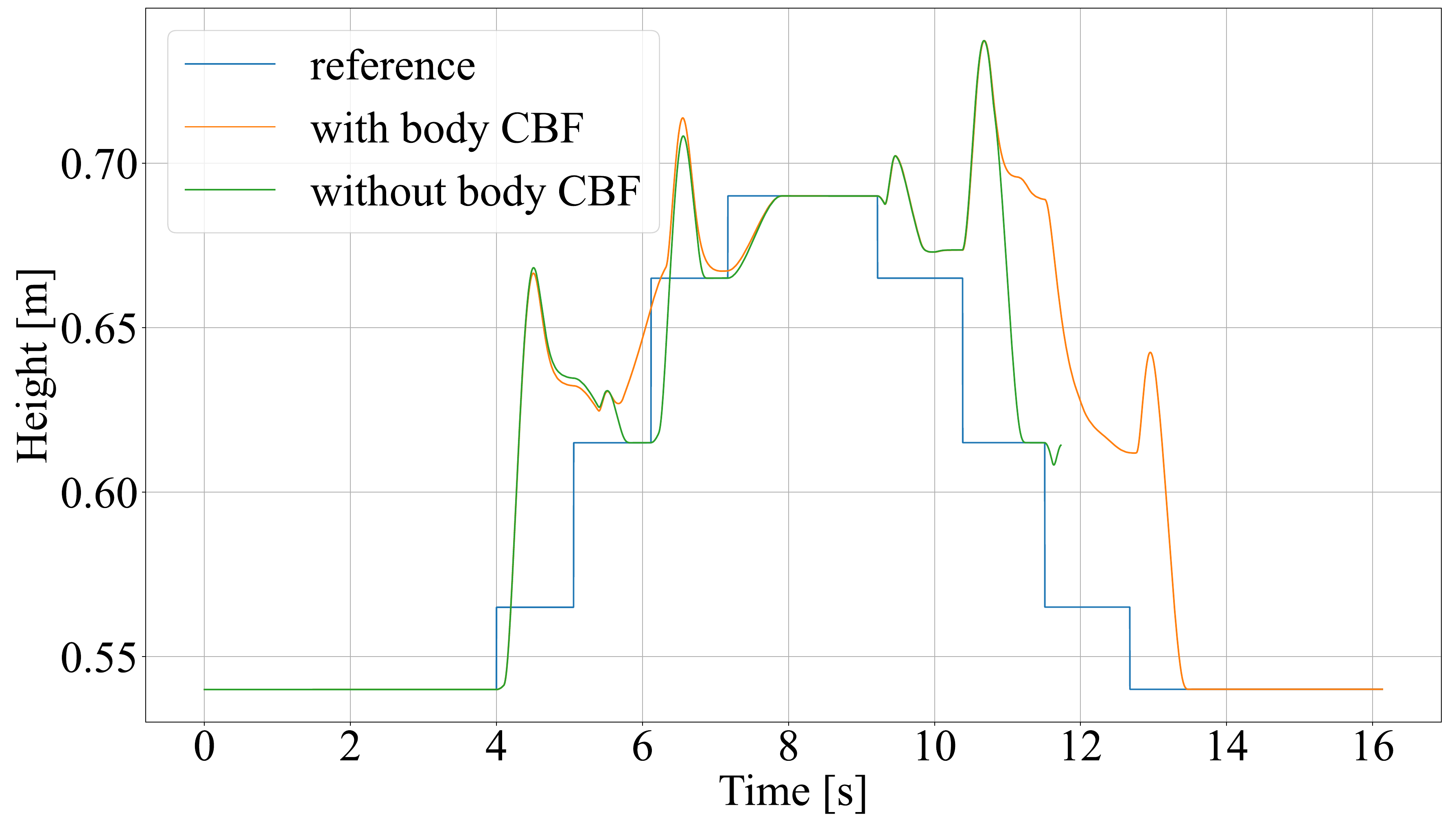} \\
     \includegraphics[scale=0.32]{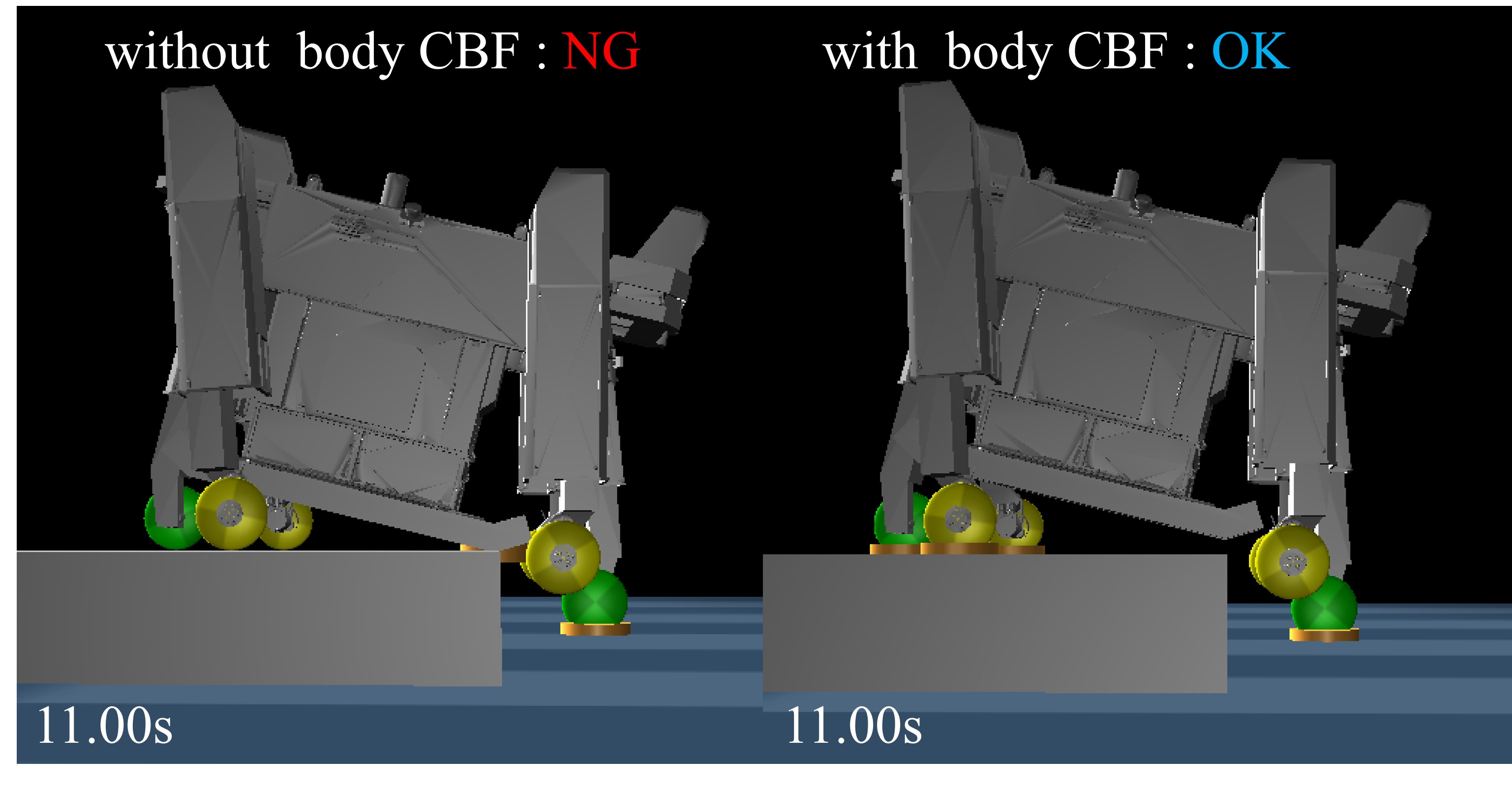}
    \caption{The above row shows height trajectories of the body. The bottom row shows the difference of the pose with and without the proposed body SSAT CBF.}
  \label{fig:sim_z_trj}
\end{figure}
\begin{figure}[tb]
    \centering
    \includegraphics[scale=0.14]{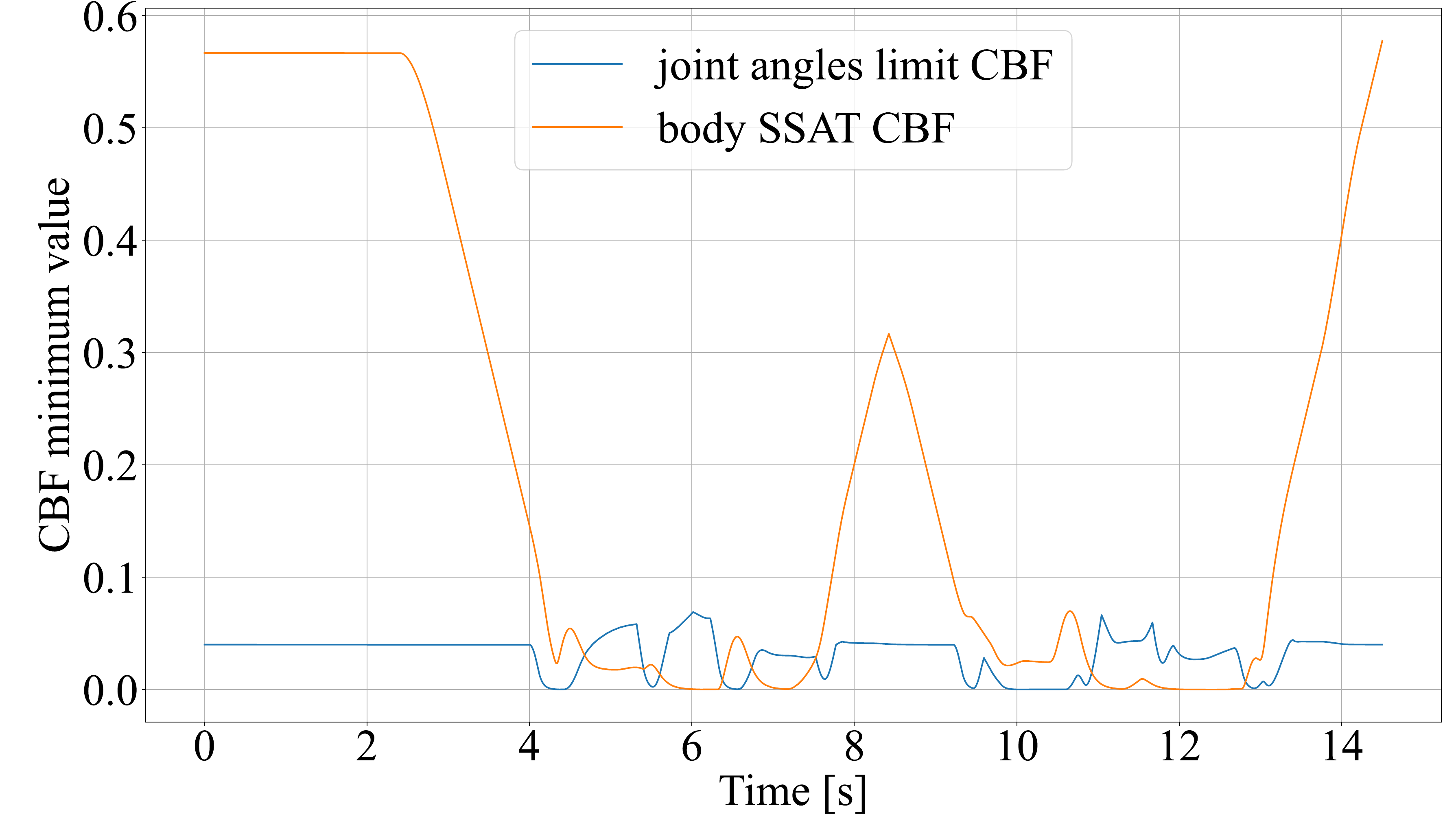} \\
    \caption{Body SSAT CBF and leg joint angles limit CBF minimum values.}
  \label{fig:sim_cbf_min_value}
\end{figure}

The CBF constraints for Tachyon 3, as proposed in Section \ref{sec:t3cbf}, have been verified by a simulation of locomotion. In this experiment, we compare the safety with and without body SSAT CBF via walking over a step using a low center of gravity posture. MuJoCo \cite{todorov2012mujoco} is used for the simulation. Fig. \ref{fig:sim_z_trj} shows the height trajectory of the body. The orange modified body trajectory is raised against the blue reference body trajectory. Notably, the green trajectory demonstrates that, without the implementation of the body SSAT CBF, the robot experiences collisions with environmental boundaries. This adverse event is effectively avoided by activating the body SSAT CBF. Furthermore, we analyzed which CBF constraints influenced the adjustment of body height. Fig. \ref{fig:sim_cbf_min_value} displays the minimum values of both the body SSAT CBF and the leg's joint angle limits CBF. The proximity of these values to zero suggests that the constraints were actively engaged. Consequently, it is evident that the cuboid collision avoidance CBF for the body and the joint angle limits CBF for the legs are efficacious in ensuring safety when the robot navigates over a step with a low center of gravity.

\subsection{Hardware Experiments for Tachyon 3}
\begin{figure}[t]
    \centering
    \includegraphics[scale=0.16]{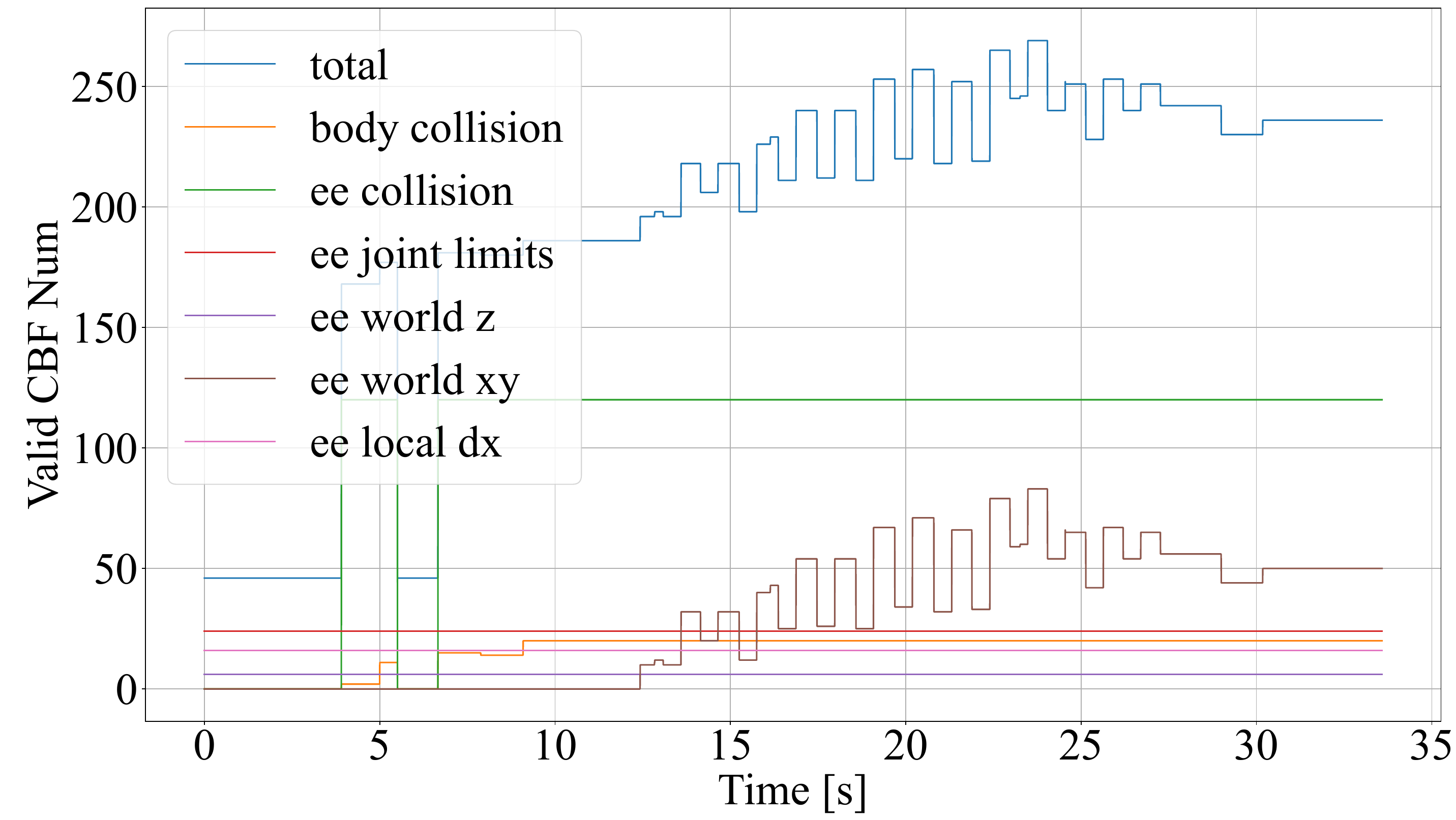} \\
    \caption{The number of valid CBF. The horizontal axis is time [s].}
  \label{fig:real_valid_cbf_num}
\end{figure}
\begin{figure}[t]
    \centering
    \includegraphics[scale=0.15]{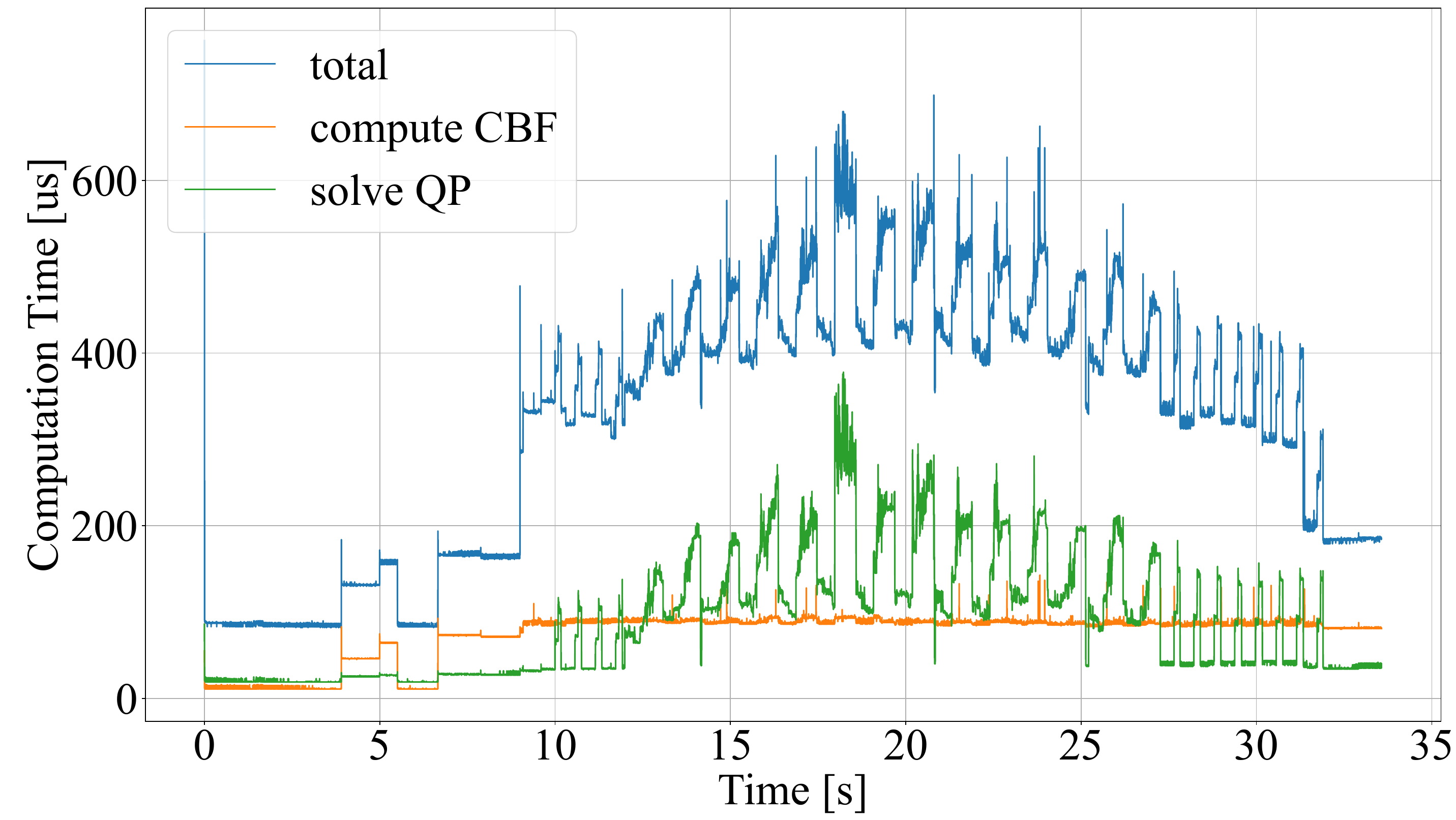} \\
    \caption{Computation time of real-time control. The vertical axis is computation time [$\mu$s] and the horizontal axis is time [s].}
  \label{fig:real_cbf_cputime}
\end{figure}
The practical usefulness of the proposed method is shown by a perceptive experiment using Tachyon 3 in an unknown environment. In this experiment, the real-time control proposed in Section \ref{sec:control} generates motion at a high frequency of 1000 Hz considering safety, such as range of motion and environmental collision avoidance, as shown in Fig. \ref{fig:intro}. Using the proposed method, Tachyon 3 walks up while recognizing 5 step stairs with a height of 16.5 cm and a depth of 35 cm. The body height is modified above by 15 cm from the standard height when walking on stairs.

Fig. \ref{fig:real_valid_cbf_num} shows the number of CBFs applied to the QP during experiments. The blue curve, representing the total number of CBFs, demonstrates that a maximum of 269 CBF constraints are computed and incorporated into the QP solver. Fig. \ref{fig:real_cbf_cputime} shows the computation time of each control update measured on the onboard PC with the CPU Intel(R) Core(TM) i7-8850 H CPU @ 2.60 GHz. The robot was able to generate motion within 1 ms as shown by the blue line (total computation time). Orange line represents the calculation time of CBF constraints with a maximum of 269 CBF constraints computed in about 80 $\mu s$. Among them, 20 obstacles are set for the six toes and the one body, resulting in a total of 140 CBF. The previous study using CBF \cite{liao2023walking} deals with only 4 obstacles of convex shape with a maximum number of vertices of 15 for 1 body's rectangle, and the paper \cite{dai2023safe} deals with 7 primitive shapes across 3 rectangular obstacles. Therefore, compared with these studies using CBF, our formulation can deal with sufficiently larger number obstacles with a sufficiently small computation time for a control cycle of 1 ms.

\section{CONCLUSIONS}\label{sec:conclusions}
In this paper, we developed a lightweight perceptive real-time control system generating feasible motion within 1 ms using an efficient kinematic model and analytically smooth CBF formulations to satisfy various constraints such as 3D collision avoidance and joint limits for Tachyon 3. Notably, we proposed a novel SSAT as an higher-order differentiable collision detection method.
The computational efficiency of SSAT was demonstrated through comparison with other collision detection methods. The selection of collision detection method depends on the required precision and the nature of the workspace, with SSAT proving particularly efficient for cuboid-approximated robots in constrained environments.
We validated the safety performance of the proposed CBF in simulations and demonstrated its practical applicability through hardware experiments of perceptive locomotion over stair climbing with on-the-fly environment perception. Our control system can generate motion without trajectory planning such as MPC. The computationally efficient SSAT CBF can be applied to body-environment collisions, other links, and self-collision avoidance.

The present system is limited to application to the movement of environments composed of planes in statically stable robots. An issue for the future is dealing with cumbersome environments and dynamic movement. Since CBF QP itself can be leveraged a real-time safety filter for strict constraint satisfaction, it is expected to work with other reference motion generation such as RL and MPC.

\bibliographystyle{IEEEtran}
\bibliography{IEEEabrv,root}

\end{document}